\documentclass[b5paper,10pt,abstractoff,DIV=calc,headings=normal,twoside]{scrartcl}
\usepackage[english]{babel}
\usepackage{lmodern}
\usepackage{microtype}
\usepackage{amsmath,amssymb,amsthm,amsfonts,graphicx,etoolbox,scrlayer-scrpage}
\usepackage[noblocks]{authblk}
\makeatletter
\patchcmd{\@maketitle}{\huge}{\Large}{}{}
\patchcmd{\abstract}{\quotation}{}{}{}
\AtBeginEnvironment{abstract}{\noindent\ignorespaces}
\AtEndEnvironment{abstract}{\par\mbox{}}
\newcommand{\shortauthor}{}
\newcommand{\shorttitle}{\@title}
\makeatother
\setkomafont{author}{\small}
\setkomafont{title}{\rmfamily\bfseries}
\setkomafont{disposition}{\rmfamily\bfseries}

\def\AMS#1{\par\noindent \textbf{AMS subject classification: }#1\par}
\newcommand{\acknowledgements}{\par\mbox{}\par\noindent\textbf{Acknowledgements: }}
\newcommand{\keywords}[1]{\par\noindent\textbf{Keywords: }#1}
\rehead[]{\shortauthor}\lohead[]{\shorttitle}\lehead[]{PPP}\rohead[]{PPP}
\ofoot[]{}\ifoot[]{}\recalctypearea
\theoremstyle{plain}

\theoremstyle{definition}

\theoremstyle{remark}

\renewenvironment{abstract}{\bigskip\noindent\begin{minipage}{\textwidth}\setlength{\parindent}{15pt}\paragraph{Abstract:}}{\end{minipage}}

\usepackage{mathtools}
\usepackage[colorlinks]{hyperref}
\usepackage{url}
\usepackage[numbers]{natbib}
\begin{document}


\renewcommand{\shortauthor}{J. Larsson}

\title{Look-Ahead Screening Rules for the Lasso}

\author[1]{Johan Larsson%
  \thanks{%
    Corresponding author: \href{mailto:johan.larsson@stat.lu.se}{\url{johan.larsson@stat.lu.se}}
  }%
}
\affil[1]{The Department of Statistics, Lund University}

\maketitle

\begin{abstract}
  The lasso is a popular method to induce shrinkage and sparsity in the
  solution vector (coefficients) of regression problems, particularly
  when there are many predictors relative to the number of observations. Solving
  the lasso in this high-dimensional setting can, however, be computationally
  demanding. Fortunately, this demand can be alleviated via the
  use of \emph{screening rules} that discard predictors prior to
  fitting the model, leading to a reduced problem to be solved.
  In this paper, we present a new screening strategy: \emph{look-ahead
    screening}. Our method uses safe screening rules to find a range of
  penalty values for which a given predictor cannot enter the model,
  thereby screening predictors along the remainder of the path.
  In experiments we show that these look-ahead screening rules outperform the
  active warm-start version of the Gap Safe rules.
\end{abstract}


\smallskip

\AMS{62J07}

\section{Introduction}\label{sec:introduction}

The lasso~\cite{tibshirani1996} is a staple among regression models for
high-dimensional data. It induces shrinkage and sparsity in the solution
vector (regression coefficients) through penalization by the
\(\ell_1\)-norm. The optimal level of penalization is, however, usually
unknown, which means we typically need to estimate it through model tuning
across a grid of candidate values: the regularization path. This leads to a
heavy computational load.

Thankfully, the advent of so-called \emph{screening rules} have lead to
remarkable advances in tackling this problem. Screening rules discard a
subset of the predictors \emph{before} fitting the model, leading to, often
considerable, reductions in problem size. There are two types of screening
rules: heuristic and safe rules.
The latter kind provides a certificate that discarded predictors cannot be
active at the optimum---that is, have a non-zero corresponding
coefficients---whereas heuristic rules do not. In this paper, we will
focus entirely on safe rules.

A prominent type of safe rules are the Gap Safe
rules~\citep{ndiaye2017,fercoq2015}, which use the duality gap in a problem
to provide effective screening rules. There currently exists
sequential versions of
the Gap Safe rules, that discard predictors for the next step on the
regularization path, as well as dynamic rules, which discard predictors
during optimization at the current penalization value.

The objective of this paper is to introduce a new screening strategy based
on Gap Safe screening: \emph{look-ahead screening}, which
screens predictors for a range of penalization parameters. We
show that this method can be used to screen predictors for the entire
stretch of the regularization path, leading to substantial improvements in
the time to fit the entire lasso path.

\section{Look-Ahead Screening}%
\label{sec:look-ahead-screening}

Let \(X \in \mathbb{R}^{n \times p}\) be the design matrix with \(n\)
observations and \(p\) predictors
and \(y \in \mathbb{R}^n\) the response vector.
The lasso is represented by the following convex optimization problem:
\begin{equation}
  \label{eq:primal}
  \operatorname*{minimize}_{\beta \in \mathbb{R}^p} \left\{P(\beta; \lambda)
  = \frac 1 2 \lVert y - X\beta \rVert_2^2 + \lambda \lVert \beta \rVert_1\right\}
\end{equation}
where \(P(\beta;\lambda)\)
is the \emph{primal} objective. We let \(\hat \beta_\lambda\) be the solution to
\eqref{eq:primal} for a given \(\lambda\).
Moreover, the dual problem of \eqref{eq:primal} is
\begin{equation}
  \label{eq:dual}
  \operatorname*{maximize}_{\theta \in \mathbb{R}^n} \left\{ D(\theta; \lambda) =
  \frac 12 y^T y - \frac{\lambda^2}{2} \left\lVert \theta
  - \frac y \lambda \right\rVert_2^2\right\}
\end{equation}
where \(D(\theta; \lambda)\) is the \emph{dual} objective.
The relationship between the primal and dual problems is given by
\(y = X\hat\beta_\lambda + \lambda \hat\theta_\lambda.\)

Next, we let \(G\) be the so-called \emph{duality gap}, defined as
\begin{equation}
  G(\beta, \theta; \lambda)
  = P(\beta; \lambda) - D(\theta; \lambda)
  =
  \frac 12 \lVert y - X\beta\rVert_2^2 + \lambda \lVert \beta \rVert_1
  - \lambda \theta^T y + \frac{\lambda^2}{2} \theta^T \theta.
\end{equation}
In the case of the lasso, strong duality holds, which means that
\(G(\hat\beta_\lambda, \hat\theta_\lambda; \lambda) = 0\) for any
choice of \(\lambda\).

Suppose, now, that we have solved the lasso for \(\lambda\); then
for any given \(\lambda^* \leq \lambda\),
the Gap Safe rule~\citep{ndiaye2017} discards the \(j\)th
predictor if
\begin{equation} \label{eq:gap-safe-rule}
  |X^T \theta_\lambda|_j + \lVert x_j\rVert_2
  \sqrt{\frac{1}{\lambda_*^2}
    G(\beta_\lambda, \theta_\lambda; \lambda^*)}
  < 1
\end{equation}
where
\[
  \theta_\lambda = \frac{y - X\beta_\lambda}{
    \max\big( |X^T(y - X\beta_\lambda)|, \lambda\big)}
\]
is a dual-feasible point~\cite{ndiaye2017} obtained through dual scaling.

Observe that \eqref{eq:gap-safe-rule} is a quadratic inequality with
respect to \(\lambda_*\), which
means that it is trivial to discover the boundary points via the
quadratic formula:
\[
  \lambda_* = \frac{-b \pm \sqrt{b^2 - 4ac}}{2a} \quad \text{where} \quad
  \begin{aligned}
    a & = \big( 1 - | x_j^T \theta_\lambda|\big)^2 -
    \frac 12 \theta_\lambda^T \theta_\lambda \lVert x_j\rVert_2^2,      \\
    b & = \big(\theta_\lambda^T y - \lVert \beta_\lambda \rVert_1\big)
    \lVert x_j \rVert_2^2,                                                          \\
    c & = - \frac 12 \lVert y - X\beta_\lambda\rVert_2^2
    \lVert x_j\rVert_2^2.                                                          \\
  \end{aligned}
\]
By restricting ourselves to an index \(j\) corresponding to a predictor that is
inactive at \(\lambda\) and recalling that we have \(\lambda_* \leq \lambda\) by
construction, we can inspect the signs of \(a\), \(b\), and \(c\) and find a
range \(\lambda\) values for which predictor \(j\) must be inactive. Using this
idea for the lasso path---a grid of \(\lambda\) values starting from the null
(intercept-only) model, which corresponds to \(\lambda_\text{max} =
\max_i|x_i^Ty|\), and finishing at fraction of this (see
\autoref{seq:simulations} for specifics)---we can screen predictor \(j\) for all
upcoming \(\lambda\)s, possibly discarding it for multiple steps on the path
rather than just the next step. We call this idea \emph{look-ahead screening}.

\begin{figure}[hbtp]
  \centering
  \includegraphics{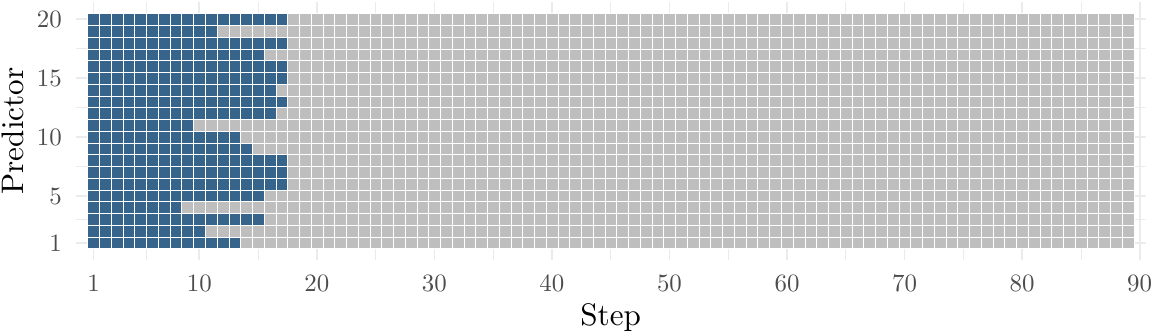}
  \caption{This figure shows the predictors screened at the first step of the
    lasso path via look-ahead screening for a random sample of
    20 predictors from the \emph{leukemia}
    data set. A blue square indicates that the corresponding predictor can
    be discarded at the respective step.\label{fig:case-study}}
\end{figure}

To illustrate the effectiveness of this screening method, we consider an
instance of employing look-ahead screening for fitting a full lasso path to the
\emph{leukemia} data set~\citep{golub1999}.  At the first step of the path, the
screening method discards 99.6\% of the predictors for the steps up to and
including step 5. The respective figures for steps 10 and 15 are 99.3\% and
57\%.  At step 20, however, the rule does not discard a single predictor.  In
\autoref{fig:case-study}, we have visualized the screening performance of
look-ahead screening for a random sample of 25 predictors from this data set.

As is typical for all screening methods, the effectiveness of look-ahead
screening is greatest at the start of the path and diminishes as the strength
of penalization decreases later on in the path. Note, however, that all of the
quantities involved in the rule are available as a by-product of solving the
problem at the previous step, which means that the costs of look-ahead
screening are diminutive.

\section{Simulations}\label{seq:simulations}

In this section, we study the effectiveness of the look-ahead screening rules by
comparing them against the active warm start version of the Gap Safe
rules~\citep{fercoq2015,ndiaye2017}.  We follow the recommendations in
\citet{ndiaye2017} and run the screening procedure every tenth pass of the
solver.
Throughout the experiments, we center the response vector by its mean, as
well as center and scale the predictors by their means and uncorrected
sample standard deviations respectively.

To construct
the regularization path, we employ the standard settings from \textsf{glmnet},
using a log-spaced path of 100 \(\lambda\) values from \(\lambda_\text{max}\)
to \(\varepsilon \lambda_\text{max}\), where \(\varepsilon = 10^{-2}\) if
\(p > n\) and \(10^{-4}\) otherwise. We also use the default
path stopping criteria from \textsf{glmnet}, that is, stop the path whenever
the deviance ratio, \(1 - \text{dev}/\text{dev}_\text{null}\),
is greater than or equal to 0.999,
the fractional increase in deviance explained is lower than
\(10^{-5}\), or, if \(p \geq n\),
when the number of active predictors exceeds or is equal to
\(n\).

To fit the lasso, we
use cyclical coordinate descent~\cite{friedman2010}.
We consider the solver to have converged whenever
the duality gap as a fraction of the primal value
for the null model is less than or equal to \(10^{-6}\) and the amount of
\emph{infeasibility}, which we define as
\(\max_j\big( |x_j^T(y - X\beta_\lambda)| -\lambda\big)\), as a
fraction of \(\lambda_\text{max}\) is lower than or
equal to \(10^{-5}\).

Source code for the experiments, including a container to facilitate
reproducibility, can be found at
\url{https://github.com/jolars/LookAheadScreening/}. An HPC
cluster node with two Intel Xeon E5-2650 v3 processors (Haswell, 20 compute
cores per node) and 64 GB of RAM was used to run the experiments.

We run experiments on a design with \(n = 100\) and \(p = 50\,000\),
drawing the rows of \(X\) i.i.d. from \(\mathcal{N}(0, \Sigma)\) and \(y\)
from \(\mathcal{N}(X\beta, \sigma^2I)\) with \(\sigma^2 =
\beta^T\Sigma\beta/\text{SNR}\), where SNR is the signal-to-noise ratio.  We set
\(5\) coefficients, equally spaced throughout the coefficient vector, to 1 and
the rest to zero. Taking inspiration from \citet{hastie2020}, we
consider SNR values of 0.1, 1, and 6.

\begin{figure}[hbtp]
  \centering
  \includegraphics{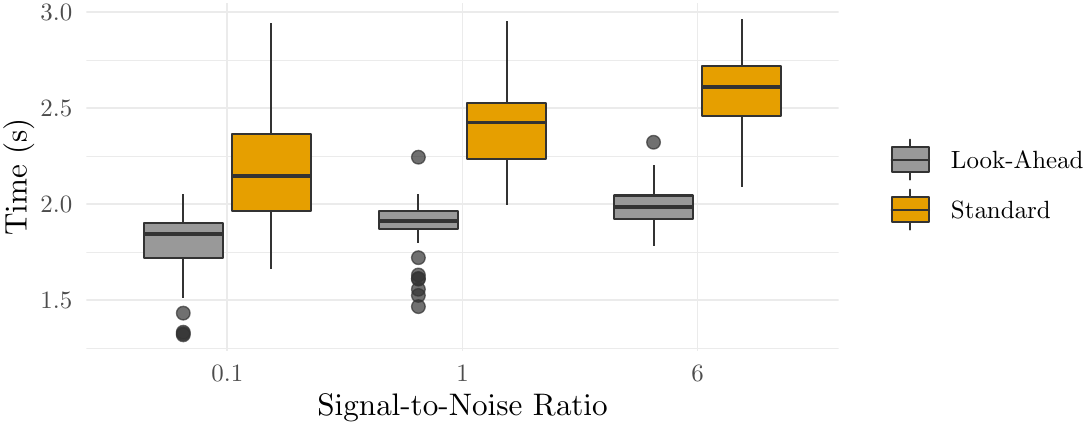}
  \caption{Standard box plots of timings to fit a full lasso path to
    a simulated data set with \(n = 100\), \(p = 50\,000\), and five true
    signals.}
  \label{fig:simulated-data}
\end{figure}

Judging by the results~(\autoref{fig:simulated-data}), the addition of
look-ahead screening results in sizable reductions in the solving time of
the lasso path, particularly in the high signal-to-noise context.

\section{Discussion}%
\label{sec:discussion}

In this paper, we have presented \emph{look-ahead screening}, which is a novel
method to screen predictors for a range of penalization values along the lasso
regularization path using Gap Safe screening. Our results show that this type of
screening can yield considerable improvements in performance for the standard
lasso. For other loss functions,
\eqref{eq:gap-safe-rule} may no longer reduce to a quadratic inequality and will
hence require more computation. Nevertheless, we believe
that applying these rules in these cases is feasible and likely to result
in comparable results.

Moreover, the idea is general and can therefore be extended to any type of safe
screening rule and also used in tandem with heuristic screening rules in order
to avoid expensive KKT computations. Finally, although we only cover one type
of cyclical
coordinate descent in our experiments, note that our screening method is
agnostic to the solver used and that we expect the results
hold for any solver that benefits from predictor screening.

\acknowledgements{
  I would like to thank my supervisor, Jonas Wallin, for valuable feedback
  on this work.

  The computations were enabled by resources provided by the Swedish
  National Infrastructure for Computing (SNIC) at LUNARC partially funded
  by the Swedish Research Council through grant agreement no. 2017-05973.
}

\bibliography{LookAheadScreening.bib}

\end{document}